\def\year{2021}\relax
\documentclass[letterpaper]{article} % DO NOT CHANGE THIS
\usepackage{aaai21}  % DO NOT CHANGE THIS
\usepackage{times}  % DO NOT CHANGE THIS
\usepackage{helvet} % DO NOT CHANGE THIS
\usepackage{courier}  % DO NOT CHANGE THIS
\usepackage[hyphens]{url}  % DO NOT CHANGE THIS
\usepackage{graphicx} % DO NOT CHANGE THIS
\usepackage{natbib}  % DO NOT CHANGE THIS AND DO NOT ADD ANY OPTIONS TO IT
\usepackage{caption} % DO NOT CHANGE THIS AND DO NOT ADD ANY OPTIONS TO IT
\usepackage{amsmath}
\usepackage{amssymb}
\usepackage{subfigure}
\usepackage{color}
\usepackage{epsfig}
\usepackage{epstopdf}
\usepackage{algorithm}
\usepackage{algorithmic}
\usepackage{multirow}
\usepackage{comment}
\usepackage{diagbox}
\usepackage{bm}

\usepackage[switch]{lineno}

\usepackage{hyperref}

\title{PC-HMR: Pose Calibration for 3D Human Mesh Recovery from 2D Images/Videos}
\author{
    Tianyu Luan \textsuperscript{\rm 1 }\thanks{T.Luan, Y.Wang and J.Zhang contributed equally.}, 
    Yali Wang \textsuperscript{\rm 1}\footnotemark[1],
	Junhao Zhang \textsuperscript{\rm 1}\footnotemark[1],
	Zhe Wang \textsuperscript{\rm 3},
	Zhipeng Zhou \textsuperscript{\rm 1},\\
	Yu Qiao \textsuperscript{\rm 1 2}\thanks{Corresponding author.}\\
}

\affiliations{
    \textsuperscript{\rm 1}ShenZhen Key Lab of Computer Vision and Pattern Recognition,\\ Shenzhen Institutes of Advanced Technology, Chinese Academy of Sciences, Shenzhen, China\\  
    \textsuperscript{\rm 2}SIAT Branch, Shenzhen Institute of Artificial Intelligence and Robotics for Society\\
    \textsuperscript{\rm 3}University of California, Irvine\\ \{ty.luan, yl.wang, zhangjh, zp.zhou, yu.qiao\}@siat.ac.cn, 
     \quad zwang15@uci.edu
}

\begin{document}

\maketitle

\begin{abstract}
The end-to-end Human Mesh Recovery (HMR) approach \cite{kanazawaHMR18} has been successfully used for 3D body reconstruction.
However,
most HMR-based frameworks reconstruct human body by directly learning mesh parameters from images or videos,
while lacking explicit guidance of 3D human pose in visual data.
As a result,
the generated mesh often exhibits incorrect pose for complex activities.
To tackle this problem,
we propose to exploit 3D pose to calibrate human mesh.
Specifically,
we develop two novel Pose Calibration frameworks,
i.e.,
Serial PC-HMR and Parallel PC-HMR.
By coupling advanced 3D pose estimators and HMR in a serial or parallel manner,
these two frameworks can effectively correct human mesh with guidance of a concise pose calibration module.
Furthermore,
since the calibration module is designed via non-rigid pose transformation,
our PC-HMR frameworks can flexibly tackle bone length variations to alleviate misplacement in the calibrated mesh.
Finally,
our frameworks are based on generic and complementary integration of data-driven learning and geometrical modeling.
Via plug-and-play modules,
they can be efficiently adapted for both image/video-based human mesh recovery.
Additionally,
they have no requirement of extra 3D pose annotations in the testing phase,
which releases inference difficulties in practice.
We perform extensive experiments on the popular benchmarks,
i.e.,
Human3.6M,
3DPW
and
SURREAL,
where
our PC-HMR frameworks achieve the SOTA results.
\end{abstract}

\section{Introduction}

3D human mesh reconstruction is an important computer vision task with wide real-life applications such as virtual try-on, robotics, etc.
However,
it is often challenging to reconstruct human body mesh from images or monocular videos,
due to inherent ambiguity in unconstrained environments.

\begin{figure}[t]
\centering
\includegraphics[width=0.95\columnwidth]{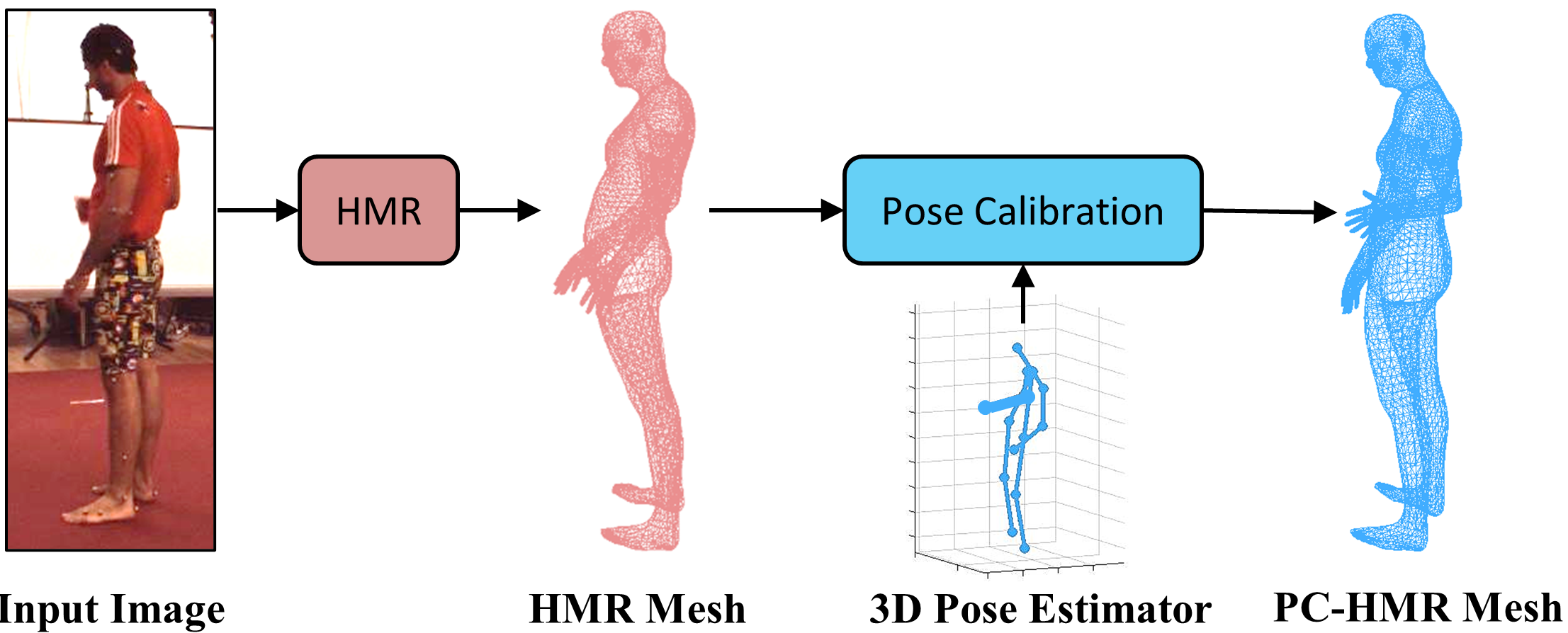}%motivation_sub.png%\includegraphics[width=0.5\textwidth]{fig/introduction.png}
\caption{Motivation.
The widely-used HMR \cite{kanazawaHMR18} reconstructs human body by directly regressing mesh parameters,
while lacking explicit guidance from 3D pose.
As a result,
it often fails to capture complex pose variations,
e.g.,
the right arm of this actor.
To tackle such problem,
we propose pose calibration for human mesh recovery (PC-HMR),
which can effectively integrate advanced 3D estimator and HMR in a novel manner to correct mesh.}
\label{fig:introduction}
\vspace{-0.3cm}
\end{figure}

Recently,
deep learning has proven to be promising to alleviate such limitation
\cite{kanazawaHMR18,learning3dhumandynamics,predicting3dhumandynamics,graphcmr,fittingintheloop,sun2019dsd-satn,cdg,kocabas2019vibe}.
In particular,
a Human Mesh Recovery (HMR) framework \cite{kanazawaHMR18} has been gradually used as the mainstream architecture for end-to-end 3D reconstruction.
However,
HMR reconstructs body configuration by directly regressing 3D mesh parameters of SMPL~\cite{SMPL:2015},
while lacking explicit and geometrical knowledge of 3D human pose.
As a result,
the generated mesh often fails to correctly capture human pose variations in complex activities,
e.g.,
HMR achieves an unsatisfactory mesh on the right arm of the actor in Fig. \ref{fig:introduction}.

Hence,
our basic idea is to use 3D human pose as guidance to reduce structural ambiguity in the 3D mesh.
However,
the ground truth 3D pose is usually unavailable in the testing phase.
Fortunately,
3D pose estimation has recently made remarkable progress \cite{simple,Cai_2019_ICCV,pavllo:videopose3d:2019},
based on the fast development of deep learning.
Inspired by this observation,
we propose to integrate 3D pose estimators and HMR approaches in a unified fashion,
so that we can take advantage of their complementary combination to calibrate human mesh.

Specifically,
we introduce three contributions in this work.
\textbf{First},
we develop two novel frameworks for Pose Calibration (PC),
i.e.,
Serial and Parallel PC-HMR.
Serial PC-HMR is a mesh-pose \textit{AutoEncoding} framework.
It can adaptively refine 3D pose to correct HMR mesh via multi-level encoding-decoding architecture.
Parallel PC-HMR is a mesh-pose \textit{TwoStream} framework.
It can directly take advantages of off-the-shelf 3D pose estimators to deform mesh in HMR.
Based on such serial or parallel manner,
these two frameworks can effectively take tradeoffs between reconstruction error and computation cost into account.
\textbf{Second},
we design a concise pose calibration module for these two frameworks.
By leveraging non-rigid pose transformation as guidance,
this module can effectively reduce misplacement caused by bone length variations,
and thus generate more natural calibrated mesh.
\textbf{Third},
our frameworks are the generic integration of pose estimators and mesh generators.
On one hand,
it can flexibly extend human mesh recovery for both images and videos.
On the other hand,
such plug-and-play design significantly releases training difficulties,
and does not need extra 3D pose annotations for inference.
We evaluate our PC-HMR frameworks on three popular benchmarks,
i.e.,
Human3.6M,
3DPW,
and SURREAL,
where
we achieve the SOTA results on mesh reconstruction.

%%%%%%%%%%%%%%%%%%%%%%%%%%%%%%%%%%%%%%%%%%%%%%%%%%%%%%%%%%%%%%%%

\section{Related Works}

\textbf{3D Mesh Reconstruction}.
SMPL~\cite{SMPL:2015} has been widely used for 3D human mesh reconstruction.
To boost its power in practice,
a number of deep learning frameworks have been proposed by using SMPL as a mesh generation module \cite{kanazawaHMR18,learning3dhumandynamics,kocabas2019vibe,sun2019dsd-satn,graphcmr,Bogo:ECCV:2016}.
In particular,
HMR \cite{kanazawaHMR18} is one mainstream framework which regresses SMPL parameters directly from input images by end-to-end training.
Following this research direction,
several extensions have been introduced to
improve 3D mesh reconstruction in images \cite{fittingintheloop,learning3dhumandynamics,sun2019dsd-satn,guler2019holopose,zeng20203d}
or model temporal relations for recovering 3D mesh in videos \cite{learning3dhumandynamics,kocabas2019vibe}.
Additionally,
several non-parametric frameworks have been proposed via voxel-based methods \cite{pifuSHNMKL19,huang2020arch}.
However,
most HMR-based approaches lack explicit guidance of 3D pose.
As a result,
the body parts of generated mesh are often located at unsatisfactory positions.
To alleviate such problem,
we leverage advanced 3D pose estimators to calibrate mesh in two general PC-HMR frameworks.

%In our frameworks, we introduce 3D human pose to mesh reconstruction to solve this problem. By using 3D pose as guidance to reconstruct a human body mesh, we manage to lower the depth uncertainty.
%Parametric human mesh reconstruction has been widely used in human body reconstruction in recent years.
%SMPL~\cite{SMPL:2015}, as the most commonly used parametric model, has been adopted in a number of following works~\cite{Guler_2019_CVPR,learning3dhumandynamics,sun2019dsd-satn,kocabas2019vibe,kolotouros2019spin,kolotouros2019cmr,Bogo:ECCV:2016,SMPL-X:2019}.
%Besides that, the development of deep learning has greatly improved this task.
%HMR~\cite{kanazawaHMR18}, as one of the first to use deep learning for human mesh reconstruction, estimates SMPL parameters directly from image features.
%On the other hand, there are also some to reconstruct human mesh. used a Although it builds a human mesh with impressive visualization, the output mesh lacks of regularization and rigging ability. \cite{} used parametric mesh model as guide for non-parametric mesh, which obtains a large improvement in mesh accuracy. Although these approaches have obtained more and more accurate human mesh, they do not properly estimate the depth of human body parts.

\textbf{3D Pose Estimation}.
Compared to 3D mesh reconstruction,
3D pose estimation has achieved more successes via deep learning.
Basically,
most current models can be categorized into two frameworks.
The first is to directly estimate 3D pose from images,
based on volumetric representation \cite{volumetric,integral,ordinal,cdg,gpa}.
But these approaches may involve in complex post-processing steps.
Based on the explosive improvement in 2D pose estimation \cite{alej2016stacked,cpn},
another framework is to estimate 2D pose from images and then lift 2D pose to 3D pose \cite{simple,semanticsgcn,Ci_2019_ICCV,Cai_2019_ICCV,pavllo:videopose3d:2019}.
Since these approaches take 2D joint locations as input,
3D human pose estimation simply focuses on learning depth of each joint.
This releases learning difficulty and leads to better 3D pose.
In this work,
we use advanced 3D pose estimators as guidance to calibrate human mesh in the HMR-based reconstruction approaches.
%On one hand,
%it solves the problem of no ground truth 3D pose in the testing phase to expand generalization ability of our frameworks.
%On the other hand,
%introducing 3D pose in HMR
This would disentangle human mesh recovery respectively into shape and pose modeling,
which effectively alleviates pose ambiguity in the generated mesh.

\section{Method}
\label{2 frameworks}

In this section,
we first introduce HMR and explain how to build up our PC-HMR frameworks.
Then,
we design a pose calibration module in our frameworks,
which uses pose transformation as guidance to correct HMR mesh.

\subsection{3D Human Mesh Recovery}

HMR \cite{kanazawaHMR18} is a widely-used deep learning framework to generate 3D human mesh from images.
First,
it uses CNN to estimate 3D mesh parameters from an input image of human,
i.e.,
$\mathbf{\Theta}=(\boldsymbol{\beta}, \boldsymbol{\theta}, \mathbf{R}, \mathbf{t}, s)=\mbox{CNN}(Img)$,
where
$\boldsymbol{\beta}$ refers to human body shape,
$\boldsymbol{\theta}$ refers to relative 3D rotation of $K$ joints,
and
$(\mathbf{R}, \mathbf{t}, s)$ represent camera parameters.
Second,
it feeds $(\boldsymbol{\beta}, \boldsymbol{\theta})$ into the well-known SMPL~\cite{SMPL:2015} to generate a triangulated mesh $\mathbf{M}^{hmr}$ with $N$ vertices,
\begin{equation}
\mathbf{M}^{hmr}=[\mathbf{V}^{hmr}_{1},\mathbf{V}^{hmr}_{2},...,\mathbf{V}^{hmr}_{N}].
\label{eq:HMRmesh}
\end{equation}
%$\mathbf{M}^{hmr}\in \mathbb{R}^{N\times3}$, where $N$ is the number of vertices.
%i.e.,
%a differentiable function that outputs a triangulated mesh,
%\begin{equation}
%\mathbf{M}^{hmr}=[\mathbf{V}^{hmr}_{1}...\mathbf{V}^{hmr}_{i}...\mathbf{V}^{hmr}_{N}],
%\label{eq:HMRmesh}
%\end{equation}
%where $N$ is the number of vertices.
%with $N = 6890$ vertices $\{\mathbf{V}^{hmr}_{i}\}_{i=1}^{N}$.
Third,
it uses a linear mesh-to-pose projector to produce 3D pose $\mathbf{J}^{hmr}$,
\begin{equation}
\mathbf{J}^{hmr}=\mathbf{U}\mathbf{M}^{hmr},
\label{eq:smpl1}
\end{equation}
where
$\mathbf{U}$ is a linear projection matrix.
%$P$ is the number of human joints after projection,
%and
%$P=14$ in HMR to match the number of joints for the LSP benchmark \cite{Johnson10}.
Finally,
it applies $(\mathbf{R}, \mathbf{t}, s)$ within a weak-perspective camera model,
which is a 3D-to-2D pose projector to obtain 2D pose $\mathbf{Z}^{hmr}$,
\begin{equation}
\mathbf{Z}^{hmr} = s\Pi(\mathbf{R}\mathbf{J}^{hmr}) + \mathbf{t},
\label{eq:smpl2}
\end{equation}
where
$\Pi$ represents the orthographic projection.
By adding supervision on ($\mathbf{\Theta}$, $\mathbf{M}^{hmr}$, $\mathbf{J}^{hmr}$, $\mathbf{Z}^{hmr}$),
HMR can be efficiently trained in an end-to-end manner.
One can refer \cite{kanazawaHMR18} for more details.

\begin{figure*}[t]
\centering
\includegraphics[width=0.78\textwidth]{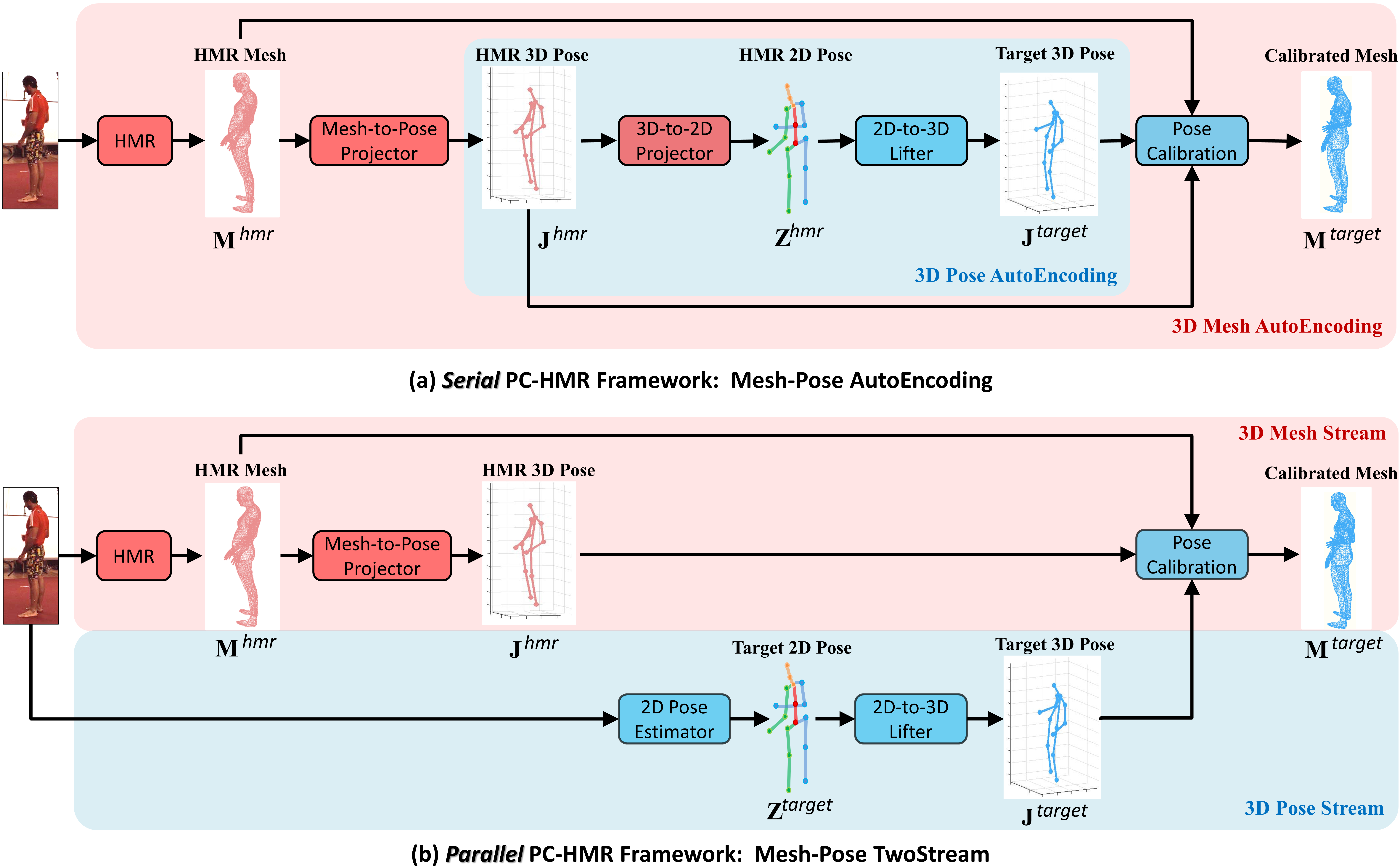}%Frame_sub1.png
\caption{PC-HMR Frameworks.
Our serial and parallel frameworks provide two generic manners to calibrate HMR mesh with explicit guidance of 3D pose.
More details can be found in Section \ref{Our PC-HMR Frameworks}.}
\label{fig:framework2}
\vspace{-0.4cm}
\end{figure*}

\subsection{Our PC-HMR Frameworks}
\label{Our PC-HMR Frameworks}

%As mentioned before,
%HMR is often limited to capture pose variations for complex activities,
%since directly regressing 3D mesh parameters $\mathbf{\Theta}$ may ignore rich geometry information of 3D pose.
%To tackle such difficulty,
%we calibrate the HMR mesh $\mathbf{M}^{hmr}$,
%with guidance of a target 3D pose $\mathbf{J}^{target}$.
%We expect that,
%$\mathbf{J}^{target}$ is the ground truth annotation.
%However,
%such annotation is usually unavailable in the testing phase.
%To address this problem,
%we design two pose calibration frameworks,
%by coupling advanced pose estimators and HMR in serial or parallel manner.

Directly regressing mesh parameters $\mathbf{\Theta}$ would introduce interference of shape modeling when learning 3D pose.
Hence,
HMR-based approaches are often limited when capturing complex pose variations.
To tackle this problem,
we design two pose calibration frameworks,
i.e.,
serial and parallel PC-HMR.
With guidance of target (or reference) 3D pose $\mathbf{J}^{target}$ from pose estimators,
serial / parallel PC-HMR can effectively calibrate HMR mesh $\mathbf{M}^{hmr}$ by mesh-pose autoencoding / twostream.

\textbf{Serial PC-HMR}.
As shown in Fig. \ref{fig:framework2} (a),
we first adapt HMR to be a multi-level autoencoding framework for calibration.
\textbf{(I) 3D Pose AutoEncoding}.
We use it to generate target 3D pose $\mathbf{J}^{target}$.
Specifically,
we notice that successful 3D pose estimators often consist of 2D pose estimator and 2D-to-3D lifter in the literature \cite{simple,pavllo:videopose3d:2019}.
Inspired by this fact,
we propose to use the HMR framework as a 2D pose estimator,
and introduce 2D-to-3D pose lifter on top of 3D-to-2D pose projector in Eq. (\ref{eq:smpl2}).
In this case,
pose projector becomes a pose encoder that encodes HMR 3D pose as HMR 2D pose ($\mathbf{J}^{hmr}\rightarrow\mathbf{Z}^{hmr}$),
and
pose lifter becomes a pose decoder that decodes HMR 2D pose as target 3D pose ($\mathbf{Z}^{hmr}\rightarrow\mathbf{J}^{target}$),
\begin{equation}
\mathbf{J}^{target} =PoseLifter(\mathbf{Z}^{hmr}).%\mbox{Lifter}
\label{eq:lifter}
\end{equation}
It is worth mentioning that,
target 3D pose $\mathbf{J}^{target}$ tends to be better than HMR 3D pose $\mathbf{J}^{hmr}$ via our autoencoding manner.
The main reason is that,
$\mathbf{J}^{hmr}$ is generated from a simple linear projection of HMR mesh $\mathbf{M}^{hmr}$ (Eq. \ref{eq:smpl1}),
which has limitation to capture pose variations as mentioned before.
On the contrary,
$\mathbf{J}^{target}$ is generated from an advanced 2D-to-3D pose lifter,
which can effectively leverage data-driven deep learning to adjust 2D locations of $\mathbf{Z}^{hmr}$ and estimate depth of each joint.
In our experiments,
we investigate several well-known pose lifters to show its effectiveness.
\textbf{(II) 3D Mesh AutoEncoding}.
After obtaining target 3D pose $\mathbf{J}^{target}$,
we design a pose calibration module to correct HMR mesh $\mathbf{M}^{hmr}$ as calibrated mesh $\mathbf{M}^{target}$.
In particular,
we add this calibration module on top of 2D-to-3D pose lifter,
leading to a 3D mesh autoencoder.
First,
we use Mesh-to-Pose projector (Eq. \ref{eq:smpl1}) as a mesh encoder,
which encodes HMR 3D mesh into HMR 3D pose ($\mathbf{M}^{hmr}\rightarrow\mathbf{J}^{hmr}$).
Then,
we use 3D pose autoencoder to map HMR 3D pose into target 3D pose ($\mathbf{J}^{hmr}\rightarrow\mathbf{J}^{target}$).
Finally,
we use the calibration module as a mesh decoder,
which deforms HMR mesh as calibrated mesh ($\mathbf{M}^{hmr}\rightarrow\mathbf{M}^{target}$),
\begin{equation}
\mathbf{M}^{target} =Calibration(\mathbf{M}^{hmr}|\mathbf{J}^{hmr}, \mathbf{J}^{target}).
\label{eq:calibrator}
\end{equation}
Note that,
our calibration function uses $(\mathbf{J}^{hmr}, \mathbf{J}^{target})$ as condition.
This is mainly because,
this function leverages non-rigid pose transformation between $\mathbf{J}^{hmr}$ and $\mathbf{J}^{target}$ as effective guidance,
to deform HMR mesh as calibrated mesh.
We will further explain this module in Section \ref{sect:calib}.
\textbf{(III) Training Serial PC-HMR}.
The output of each encoder and decoder in our serial PC-HMR has its physical meaning such as 3D mesh, 3D pose or 2D pose.
This makes our training procedure become convenient and flexible,
i.e.,
we can train each module of our serial PC-HMR separately,
and then fine-tune the entire framework.
In our experiments,
we first pretrain HMR (including mesh-to-pose and 3D-to-2D projectors) and 2D-to-3D pose lifter separately.
Then,
we fine-tune the entire framework,
by adding 3D pose supervision on pose lifter and 3D mesh supervision on pose calibration module. %
As a result,
our serial PC-HMR can effectively calibrate HMR mesh by 3D pose refinement.

\textbf{Parallel PC-HMR}.
To obtain better target 3D pose for calibration,
we adapt HMR as a twostream framework in Fig. \ref{fig:framework2} (b).
\textbf{(I) 3D Pose Stream}.
Instead of pose refinement in serial PC-HMR,
we directly use advanced 3D pose estimator as extra stream to generate target 3D pose $\mathbf{J}^{target}$,
e.g.,
we first apply 2D pose estimator to obtain 2D pose from the input image ($Img\rightarrow\mathbf{Z}^{target}$),
and then use 2D-to-3D pose lifter to estimate target 3D pose from 2D pose ($\mathbf{Z}^{target}\rightarrow\mathbf{J}^{target}$).
Note that,
2D pose $\mathbf{Z}^{target}$ is better than $\mathbf{Z}^{hmr}$ in serial PC-HMR.
The main reason is that,
$\mathbf{Z}^{hmr}$ is indirectly generated through HMR, Mesh-to-Pose projector, 3D-to-2D projector with complex interference of shape modeling.
On the contrary,
$\mathbf{Z}^{target}$ is directly generated from advanced 2D pose estimator,
which is able to predict 2D joint locations accurately from an input image.
Subsequently,
by using a better 2D pose $\mathbf{Z}^{target}$,
target 3D pose $\mathbf{J}^{target}$ from this extra stream tends to be more reliable than the one (Eq. \ref{eq:lifter}) from serial PC-HMR.
%\textcolor{red}{Note that the sole purpose of 3D Pose Stream is to provide the framework with more accurate 3D pose, so in practice is can be replaced by any 3D pose generator (i.e. one-stage 3D pose generator).}
\textbf{(II) 3D Mesh Stream}.
After obtaining target 3D pose,
we introduce a 3D mesh stream,
where
we first obtain human mesh from HMR,
and then correct it via pose calibration module (Eq. \ref{eq:calibrator}).
\textbf{(III) Training Parallel PC-HMR}.
Similar to serial PC-HMR,
parallel PC-HMR can be trained in a flexible manner.
In our experiments,
we first pretrain mesh stream (HMR+Mesh-to-Pose Projector) and pose stream
%(2D Estimator+2D-to-3D Lifter)
separately,
and then fine-tune the entire framework by adding 3D pose supervision on pose lifter and 3D mesh supervision on calibration module.

\textbf{Discussions}.
We discuss the proposed PC-HMR frameworks from the following aspects.
\textbf{(I) Extension on Video-Based Mesh Recovery}.
Our PC-HMR frameworks provide two generic manners to reconstruct human mesh with explicit guidance of 3D pose.
Hence,
they can be straightforwardly extended for video-based mesh reconstruction,
by integrating video-based mesh generators and pose estimators in our plug-and-play fashion.
In our experiments,
we will show flexibility and robustness of our PC-HMR frameworks for both image/video-based mesh recovery.
\textbf{(II) Accuracy-Efficiency Tradeoffs}.
Our PC-HMR frameworks take tradeoffs between reconstruction error and computation cost into account,
i.e.,
parallel framework achieves a smaller reconstruction error with extra 3D pose estimation stream,
while
serial framework maintains a lighter computation cost by 3D pose refinement with 2D-to-3D pose lifter.
In practice,
both can effectively calibrate HMR mesh.
We can choose either of them,
depending on accuracy-efficiency balance.

\subsection{Calibration Module}
\label{sect:calib}

%In this section,
%we further explain the pose calibration module (Eq. \ref{eq:calibrator}) in our PC-HMR frameworks.
%Specifically,
%we first establish a concise non-rigid transformation between HMR 3D pose $\mathbf{J}^{hmr}$ and target 3D pose $\mathbf{J}^{target}$.
%Then,
%we use this pose transformation as reference to deform HMR mesh $\mathbf{M}^{hmr}$ into our calibrated mesh $\mathbf{M}^{target}$.

In this section,
we further explain pose calibration module (Eq. \ref{eq:calibrator}) in PC-HMR frameworks.
To leverage target 3D pose $\mathbf{J}^{target}$ as guidance,
we propose to establish transformation between $\mathbf{J}^{target}$ and HMR 3D pose $\mathbf{J}^{hmr}$.
Then,
we use this pose transformation as reference to deform HMR mesh $\mathbf{M}^{hmr}$ into our calibrated mesh $\mathbf{M}^{target}$.
Note that,
since $\mathbf{J}^{hmr}$ and $\mathbf{J}^{target}$ may not share the same bone lengths,
our transformation is designed to be non-rigid to alleviate misplacement in the calibrated mesh.
Moreover,
our module is based on geometrical transformation with learnable parameters.
Hence,
it can take advantages of both physical and data-driven learning to calibrate HMR mesh effectively.

\begin{figure}[t]
\centering
\includegraphics[width=0.95\columnwidth]{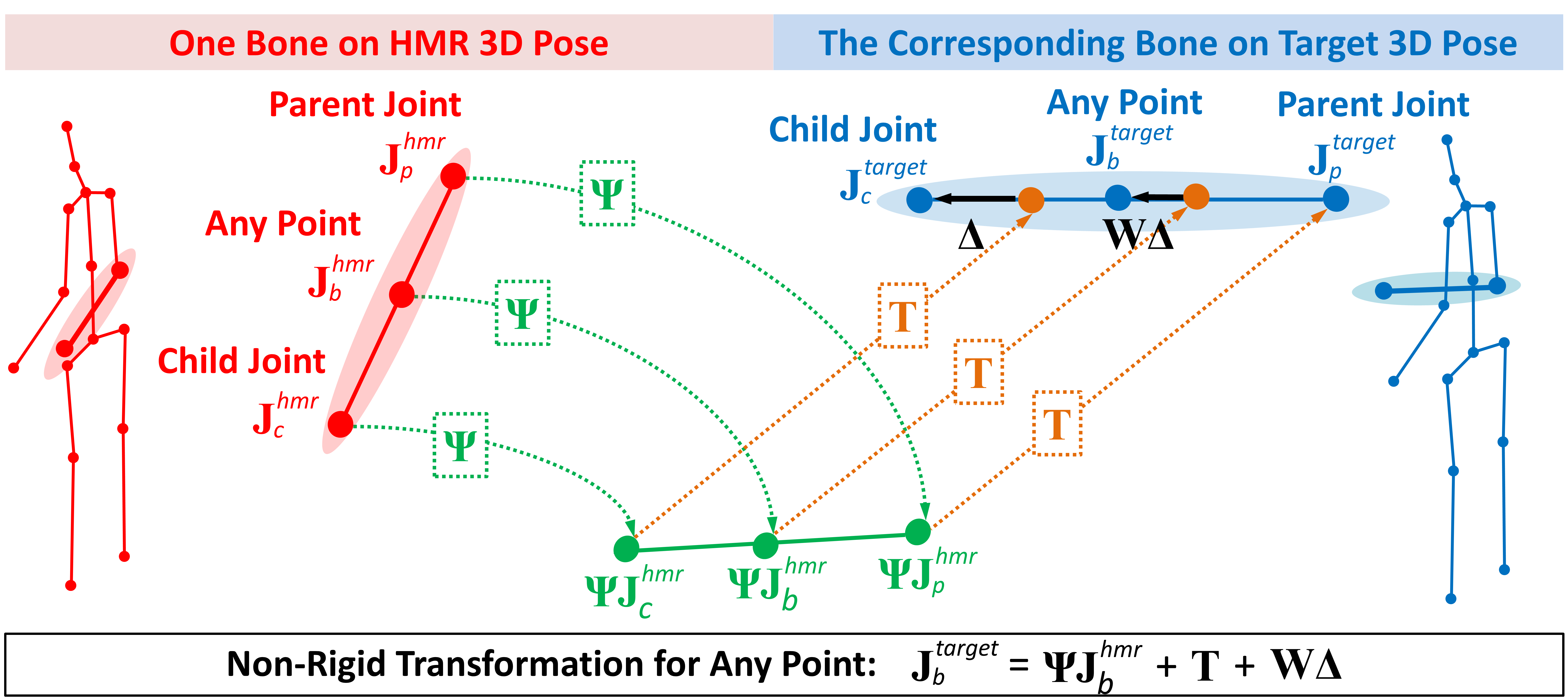}%transform_sub.png
\caption{Non-Rigid Pose Transformation.
For an arbitrary point $\mathbf{J}^{hmr}_{b}$ on a bone of HMR 3D pose,
we design a non-rigid transformation to find its corresponding point $\mathbf{J}^{target}_{b}$ on the corresponding bone of target 3D pose.
Note that,
the bone length of target 3D pose can be different from that of HMR 3D pose.
Our non-rigid transformation can effectively tackle this problem by bone extension or shortening (i.e., $\mathbf{W}\bm{\Delta}$).
}%
\label{fig:deformer}
\vspace{-0.4cm}
\end{figure}

\textbf{Non-Rigid Pose Transformation}.
%For calibrating HMR mesh according to target 3D pose,
%we have to find 3D pose transformation between $\mathbf{J}^{hmr}$ and $\mathbf{J}^{target}$.
%Note that,
%since $\mathbf{J}^{hmr}$ and $\mathbf{J}^{target}$ may not share the same bone lengths,
%we consider that the transformation would be non-rigid with bone extension or shortening.
Without loss of generality,
we mainly describe how to make non-rigid transformation on one bone.
Specifically,
for one bone in the HMR 3D pose $\mathbf{J}^{hmr}$,
we denote
($\mathbf{J}^{hmr}_{p}$,
$\mathbf{J}^{hmr}_{c}$,
$\mathbf{J}^{hmr}_{b}$)
respectively as
the parent joint of this bone,
the child joint of this bone,
and
an arbitrary point on this bone between parent and child joints.
For the corresponding bone in the target 3D pose $\mathbf{J}^{target}$,
we use the similar notations such as
($\mathbf{J}^{target}_{p}$,
$\mathbf{J}^{target}_{c}$,
$\mathbf{J}^{target}_{b}$).
The parent and child joints in the bone are defined according to human skeleton,
i.e.,
the top joint in the bone is parent,
and the bottom joint in the bone is child.
Note that,
the parent and child joints are given in both HMR and target 3D poses.
Hence,
\textit{our goal is to build up non-rigid transformation of any point between $\mathbf{J}^{hmr}_{b}$ and $\mathbf{J}^{target}_{b}$},
based on the parent and child joints ($\mathbf{J}^{hmr}_{p}$, $\mathbf{J}^{hmr}_{c}$) and ($\mathbf{J}^{target}_{p}$, $\mathbf{J}^{target}_{c}$).
The whole transformation process is shown in Fig. \ref{fig:deformer}.
\textbf{(I) Rotation $\bm{\Psi}$}.
We first compute the rotation matrix $\bm{\Psi}$,
in order to rotate the bone in HMR pose along the direction of the corresponding bone in target pose.
Specifically,
based on Lie algebra,
we can compute the rotation vector $\bm{\psi}$ between bone directions $\bm{b}^{hmr}$ and $\bm{b}^{target}$,
\begin{equation}
\bm{\psi}=arccos\left(\frac{\bm{b}^{hmr} \bm{b}^{target}}{\|\bm{b}^{hmr}\Vert\|\bm{b}^{target}\Vert}\right)\frac{\bm{b}^{hmr} \times\bm{b}^{target}}{\|\bm{b}^{hmr} \times \bm{b}^{target}\Vert},
\label{eq:bone1}
\end{equation}
where
$\bm{b}^{hmr}=\mathbf{J}^{hmr}_{p}-\mathbf{J}^{hmr}_{c}$,
$\bm{b}^{target}=\mathbf{J}^{target}_{p}-\mathbf{J}^{target}_{c}$
and
$\times$ is cross product.
Then,
% we use Rodrigues Transformation of $\bm{\psi}$ to obtain the rotation matrix $\bm{\Psi}$,
we use Rodrigues rotation formula \cite{koks2006explorations} to transform rotation vector $\bm{\psi}$ into rotation matrix $\bm{\Psi}$,
\begin{equation}
\boldsymbol{\Psi} =  cos\|\bm{\psi}\Vert \mathbf{I} + (1-cos\|\bm{\psi}\Vert) \bm{\phi} \bm{\phi}^{T} + sin \|\bm{\psi}\Vert \bm{\phi}^{\bigwedge},
%Rodrigues(\bm{\psi}).
\label{eq:bone_trans02}
\end{equation}
where $\bm{\phi} = \frac{\bm{\psi}}{\|\bm{\psi}\Vert}$ is the unit vector of $\bm{\psi}$,
$\bm{\phi}^{T}$ is transpose of $\bm{\phi}$,
and
$\bm{\phi}^{\bigwedge} = \left[ \begin{array}{ccc}
0 & -\bm{\phi}_{z} & \bm{\phi}_{y}\\
\bm{\phi}_{z} & 0 & -\bm{\phi}_{x}\\
-\bm{\phi}_{y} & \bm{\phi}_{x} & 0
\end{array} 
\right ]$ is the cross product matrix of $\bm{\phi}$.

\textbf{(II) Translation $\mathbf{T}$}.
After joint rotation,
we need to align the rotated joint of HMR bone with the corresponding joint of target bone.
Specifically,
we use parent joint as reference,
and compute the translation vector $\mathbf{T}$ for alignment,
\begin{equation}
\mathbf{T} = \mathbf{J}^{target}_{p} - \boldsymbol{\Psi}\mathbf{J}^{hmr}_{p}.
\label{eq:bone_trans1}
\end{equation}
\textbf{(III) Non-Rigid Term $\bm{\Delta}$}.
Given rotation and translation,
we next transform the child joint of HMR bone $\mathbf{J}^{hmr}_{c}$ into the space of target bone,
\begin{equation}
\breve{\mathbf{J}}^{target}_{c} = \boldsymbol{\Psi}\mathbf{J}^{hmr}_{c} + \mathbf{T}.
\label{eq:bone_trans2}
\end{equation}
However,
there may be gap between $\breve{\mathbf{J}}^{target}_{c}$ and the given child joint of target pose $\mathbf{J}^{target}_{c}$,
due to bone length variations between HMR and target poses.
To fill up this gap,
we propose a non-rigid term on the child joint,
\begin{equation}
\bm{\Delta} = \mathbf{J}^{target}_{c} - \breve{\mathbf{J}}^{target}_{c}.
\label{eq:bone_trans30}
\end{equation}
As a result,
for an arbitrary point on the bone of HMR pose $\mathbf{J}^{hmr}_{b}$,
we can find its corresponding point on the target pose $\mathbf{J}^{target}_{b}$,
according to a non-rigid transformation as follows,
\begin{equation}
\mathbf{J}^{target}_{b}=\boldsymbol{\Psi}\mathbf{J}^{hmr}_{b} + \mathbf{T}+ \mathbf{W}\bm{\Delta}.
\label{eq:bone_trans3}
\end{equation}
As shown in Eq. (\ref{eq:bone_trans3}),
we first transform $\mathbf{J}^{hmr}_{b}$ into the space of target pose,
i.e.,
$\breve{\mathbf{J}}^{target}_{b}=\boldsymbol{\Psi}\mathbf{J}^{hmr}_{b} + \mathbf{T}$.
Then,
we further improve $\breve{\mathbf{J}}^{target}_{b}$ by proportionally adjusting non-rigid term $\mathbf{W}\bm{\Delta}$ with a learnable parameter matrix $\mathbf{W}$.
In this case,
our non-rigid transformation can extend or shorten $\breve{\mathbf{J}}^{target}_{b}$ along the target bone,
in order to make correct alignment between HMR and target 3D poses.

\begin{figure}[t]
\centering
\includegraphics[width=0.90\columnwidth]{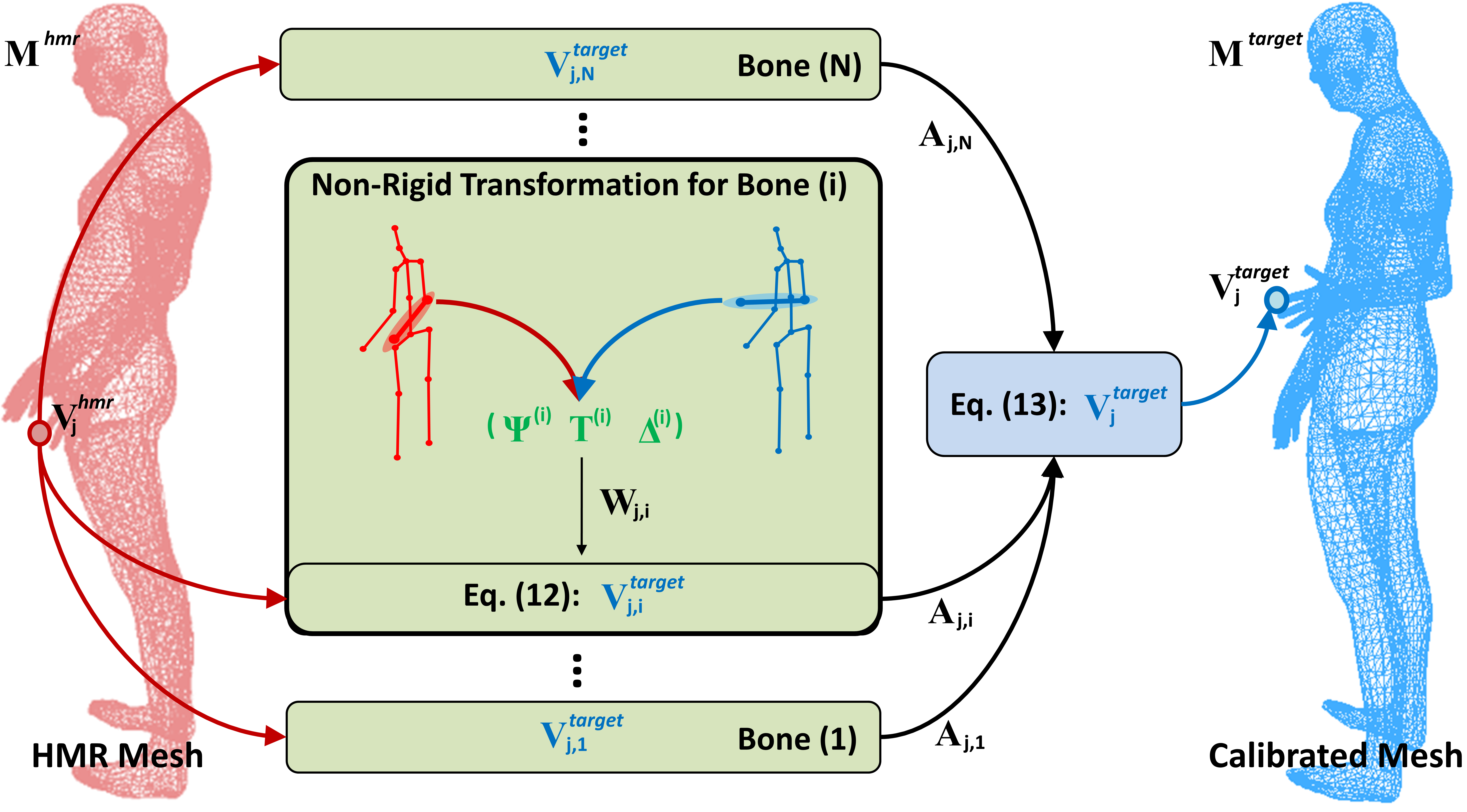}%Deform_sub.png
\caption{Mesh calibration with guidance of non-rigid 3D pose transformation.
For the $j$-th vertex $\mathbf{V}_{j}^{hmr}$ in the HMR mesh,
we first transform it as $\mathbf{V}^{target}_{j,i}$ in Eq. (\ref{eq:mesh_trans2}),
according to pose transformation of the $i$-th bone $(\boldsymbol{\Psi}^{(i)}, \mathbf{T}^{(i)}, \bm{\Delta}^{(i)})$.
Then,
we summarize the contribution of all the bones by a learnable weight matrix $\mathbf{A}$,
and produce the final prediction of the $j$-th vertex in the target mesh by Eq. (\ref{eq:mesh_trans3}).}
\label{fig:deformer1}
\vspace{-0.4cm}
\end{figure}

\textbf{Mesh Calibration}.
After obtaining pose transformation in Eq. (\ref{eq:bone_trans3}),
we use it as an distinct guidance to calibrate HMR mesh.
Specifically,
we assume that each vertex on the mesh follows the similar transformation in Eq. (\ref{eq:bone_trans3}).
In this case,
we can transform the $j$-th vertex $\mathbf{V}_{j}^{hmr}$ in the HMR mesh $\mathbf{M}^{hmr}$ as
\begin{equation}
\mathbf{V}^{target}_{j,i}=\boldsymbol{\Psi}^{(i)}\mathbf{V}^{hmr}_{j} + \mathbf{T}^{(i)}+ \mathbf{W}_{j,i}\bm{\Delta}^{(i)},
\label{eq:mesh_trans2}
\end{equation}
where
$\mathbf{V}^{target}_{j,i}$ is the prediction of the corresponding vertex in the target mesh, %$\mathbf{M}^{target}$
according to pose transformation of the $i$-th bone $(\boldsymbol{\Psi}^{(i)}, \mathbf{T}^{(i)}, \bm{\Delta}^{(i)})$ .
Subsequently,
we summarize the contribution of all the bones by a learnable weight matrix $\mathbf{A}$,
and produce the final prediction of the $j$-th vertex in the target mesh $\mathbf{M}^{target}$,
\begin{equation}
\mathbf{V}^{target}_{j}=\sum\nolimits_{i}\mathbf{A}_{j,i}\mathbf{V}^{target}_{j,i}. %+ pbs(\delta\theta(\boldsymbol{\Psi}^{(i)}, \theta)).
\label{eq:mesh_trans3}
\end{equation}
We can see in Eq. (\ref{eq:mesh_trans2})-(\ref{eq:mesh_trans3}) that,
$\mathbf{V}^{target}_{j}$ is generated by non-rigid pose transformation with learnable parameters $\mathbf{W}_{j,i}$ and $\mathbf{A}_{j,i}$.
Hence,
our calibration module can integrate geometrical modeling and data-driven learning together to effectively calibrate HMR mesh.

\begin{table}[t]
\begin{center}
\resizebox{0.47\textwidth}{!}{
\renewcommand\tabcolsep{3pt}
\begin{tabular}{l|c c c}
\hline\hline
Human3.6M
&MPJPE$\downarrow$ &PA-MPJPE$\downarrow$ &MPVE$\downarrow$ \\
\hline
	    Self-mocap~\cite{selfmocap}
	    &98.4 &- &145.8\\
%	    Bodynet~\cite{bodynet}
%	    &49.0 &- &-\\
	    HMR~\cite{kanazawaHMR18}
	    &88.0 &56.8 &96.1\\
	    HMMR~\cite{learning3dhumandynamics}
	    &85.2 &56.7 &94.2\\
	    TemporalContext~\cite{ArnabCVPR2019}
	    &77.8  &54.3  &- \\
	    GraphCMR~\cite{KolotourosCVPR2019}
	    &- &50.1 &-\\
	    VIBE ~\cite{kocabas2019vibe}
	    &65.9 &41.5 &-\\
	    Pose2Mesh ~\cite{ChoiECCV2020}
	    &64.9 &47.0  &-\\
	    HoloPose ~\cite{guler2019holopose}
	    &60.3 &46.5 &-\\
	    HKMR ~\cite{GeorgakisECCV2020}
	    &59.6 &- &-\\
	    DSD-SATN ~\cite{sun2019dsd-satn}
	    &59.1 &42.4 &-\\
	    I2L-MeshNet ~\cite{MoonECCV2020}
	    &55.7 &41.7 &-\\
	    DaNet ~\cite{zhang2019danet}
	    &54.6 &42.9 &66.5\\
	    Occluded ~\cite{ZhangCVPR2020}
	    &- &41.7 &-\\
	    SPIN ~\cite{fittingintheloop}
	    &-&41.1&-\\
	    DecoMR ~\cite{zeng20203d}
	    &- &39.3 &-\\
%	    Cross-dataset ~\cite{cdg}
%	    &52.0 &42.5 &-\\
        \textit{\textbf{Our PC-HMR}}
        &\textbf{47.9} &\textbf{37.3} &\textbf{61.1}\\
\hline\hline
3DPW
&MPJPE$\downarrow$ &PA-MPJPE$\downarrow$ &MPVE$\downarrow$ \\
\hline
	    HMR~\cite{kanazawaHMR18}
	    &128.4 &81.8 &152.7\\
%	    Occluded ~\cite{ZhangCVPR2020}
%	    &- &72.2 &-\\
%	    GraphCMR~\cite{KolotourosCVPR2019}
%	    &- &70.2 &-\\
	    DSD-SATN ~\cite{sun2019dsd-satn}
	    &122.7 &69.5 &183.4\\
	    SPIN ~\cite{fittingintheloop}
	    &96.9&59.2&116.4\\
	    VIBE ~\cite{kocabas2019vibe}
	    &93.5 &\textbf{56.5} &113.4\\
	    I2L-MeshNet ~\cite{MoonECCV2020}
	    &93.2 &58.6 &-\\	
	    Pose2Mesh ~\cite{ChoiECCV2020}
	    &89.2 &58.9  &-\\
%	    Cross-dataset ~\cite{cdg}
%	    &89.7 &65.2 &-\\
	    \textit{\textbf{Our PC-HMR}}
	    &\textbf{87.8} &66.9 &\textbf{108.6}\\
\hline\hline
SURREAL
&MPJPE$\downarrow$ &PA-MPJPE$\downarrow$ &MPVE$\downarrow$ \\
\hline
	    HMR~\cite{kanazawaHMR18}
	    &73.6 &55.4 &85.1\\
	    Self-mocap~\cite{selfmocap}
	    &64.4 &- &74.5 \\	
	    Bodynet~\cite{bodynet}
	    &- &- &73.6 \\
%	    Cross-dataset ~\cite{cdg}
%	    &37.1 &31.7 &- \\
	   % DecoMR ~\cite{zeng20203d}
	   % &- &- &56.5 \\
	   \textit{\textbf{Our PC-HMR}}
	    &\textbf{51.7} &\textbf{37.9} &\textbf{59.8} \\
\hline\hline
\end{tabular}
}
\end{center}
\caption{SOTA comparison for human mesh reconstruction on Human3.6M, 3DPW and SURREAL datasets.
For most metrics and benchmarks,
our parallel PC-HMR framework achieves the SOTA performance,
e.g.,
for Human3.6M,
it outperforms \cite{sun2019dsd-satn,ChoiECCV2020} that also leverage pose estimators for mesh reconstruction.
This shows our PC-HMR framework is a more effective manner to boost mesh recovery by human pose.}
\label{tab:SOTA}
\vspace{-0.4cm}
\end{table}

\begin{table*}[t]
\begin{center}
\resizebox{0.88\textwidth}{!}{
\renewcommand\tabcolsep{3pt}
\begin{tabular}{c||c|c||c|c||c|c||c|c}
\hline\hline
\multirow{3}{*}{Designs}
& \multicolumn{2}{c||}{Mesh Generator}
& \multicolumn{2}{c||}{2D-to-3D Pose Lifter}
& \multicolumn{2}{c||}{2D Pose Estimator}
& \multirow{3}{*}{MPVE$\downarrow$}
& \multirow{3}{*}{GFLOPs$\downarrow$}
\\
\cline{2-7}
& HMR
& HMMR
& SemanticGCN
& VideoPose3D
& CPN
& HRNet
&
&
\\
& \cite{kanazawaHMR18}
& \cite{learning3dhumandynamics}
& \cite{semanticsgcn}
& \cite{pavllo:videopose3d:2019}
& \cite{hong2018cascaded}
& \cite{SunCVPR2019}
&
&
\\
\hline\hline

\multirow{2}{*}{Baseline}
& $\checkmark$
& -
& -
& -
& -
& -
&96.1
&3.5
\\ %\cline{2-9}

& -
& $\checkmark$
& -
& -
& -
& -
&94.2
&3.6
\\ \hline

\multirow{4}{*}{Serial}
& $\checkmark$
& -
& $\checkmark$
& -
& -
& -
&91.9
&3.5
\\ %\cline{2-9}

& $\checkmark$
& -
& -
& $\checkmark$
& -
& -
&82.7
&3.5
\\ %\cline{2-9}

& -
& $\checkmark$
& $\checkmark$
& -
& -
& -
&90.3
&3.6
\\ %\cline{2-9}

& -
& $\checkmark$
& -
& $\checkmark$
& -
& -
&82.7
&3.6
\\\hline

\multirow{8}{*}{Parallel}
& $\checkmark$
& -
& $\checkmark$
& -
& $\checkmark$
& -
&79.0
&16.1
\\ %\cline{2-9}

& $\checkmark$
& -
& $\checkmark$
& -
& -
& $\checkmark$
&77.3
&19.5
\\ %\cline{2-9}

& $\checkmark$
& -
& -
& $\checkmark$
& $\checkmark$
& -
&61.1
&16.1
\\ %\cline{2-9}

& $\checkmark$
& -
& -
& $\checkmark$
& -
& $\checkmark$
&68.6
&19.5
\\ %\cline{2-9}

& -
& $\checkmark$
& $\checkmark$
& -
& $\checkmark$
& -
&77.1
&16.2
\\ %\cline{2-9}

& -
& $\checkmark$
& $\checkmark$
& -
& -
& $\checkmark$
&76.2
&19.6
\\ %\cline{2-9}

& -
& $\checkmark$
& -
& $\checkmark$
& $\checkmark$
& -
&61.1
&16.2
\\ %\cline{2-9}

& -
& $\checkmark$
& -
& $\checkmark$
& -
& $\checkmark$
&68.4
&19.6
\\
\hline\hline
\end{tabular}	
}
\end{center}
\caption{Generalization capacity (Human3.6M).
Our frameworks can be straightforwardly used for video-based mesh reconstruction without any difficulties,
e.g.,
we apply video-based HMMR \cite{learning3dhumandynamics} as mesh generator,
or apply video-based VideoPose3D \cite{pavllo:videopose3d:2019} as 2D-to-3D pose lifter.
More explanations can be found in Section \ref{sect:ablation}.
}
%First,
%our frameworks can be straightforwardly used for video-based mesh reconstruction without any difficulties,
%e.g.,
%in the serial framework,
%we use HMMR \cite{humanMotionKZFM19} as video-based mesh generator,
%and VideoPose3D \cite{pavllo:videopose3d:2019} as video-based 2D-to-3D pose lifter.
%Second,
%serial PC-HMR is more lighter while parallel PC-HMR is more accurate.
%This provides more flexible choices in practice,
%depending on which factor (recovery error or computation cost) is important in the deployment.
%%%%%%%%%%%%%%%%%%
%We test that both our frameworks are built in plug-in style for mesh reconstruction, pose lifter and 2D pose estimator (only needed in Two-stream framework) modules. We report the MPVE and GFLOPs of every combination of chosen methods. For every number, lower is better. The underline ones indicate approaches of video inputs. Others are single-frame inputs methods.
\label{tab:plugin}
\vspace{0.1cm}
%\end{table*}
%
%
%\begin{table*}[t]
\begin{center}
\resizebox{0.9\textwidth}{!}{
\begin{tabular}{l||c||c|cccccc}
  \hline\hline
  Method & HMR & Our PC-HMR & w/o Non-Rigid & w/o Fine-Tune & PC-Template & GT 3D Pose & Extra Self-Rotation & w/o Shape Compensation \\
  \hline
  MPVE   & 96.1    &    61.1        &     69.3          &     85.5       &    125.72         &      29.9      &  61.0 & 61.2 \\
  \hline\hline
\end{tabular}
%\begin{tabular}{l|c c c}
%\hline\hline
%HMR		             &MPJPE$\downarrow$ &PA-MPJPE$\downarrow$ &MPVE$\downarrow$ \\
%\hline
%Baseline             &           &      &         \\
%\hline\hline
%Our PC-HMR		     &MPJPE$\downarrow$ &PA-MPJPE$\downarrow$ &MPVE$\downarrow$ \\
%\hline
%Rigid                &           &      &         \\
%Non-Rigid            &           &      &         \\
%\hline
%w/o Fine-Tune        &           &      &         \\
%Fine-Tune            &           &      &         \\
%\hline
%PC-Template          &           &      &         \\
%PC-HMR               &           &      &         \\
%\hline
%Estimated 3D Pose    &           &      &         \\
%GT 3D Pose           &           &      &         \\
%\hline
%Default              &           &      &         \\
%Extra Self-Rotation  &           &      &         \\
%\hline\hline
%\end{tabular}
}
\end{center}
\caption{Detailed designs of our PC-HMR (Human3.6M).}
\label{tab:calib}
\vspace{-0.5cm}
\end{table*}

\section{Experiments}

% \textcolor{red}{
\textbf{Datasets and Implementation Details}.
To evaluate our PC-HMR frameworks,
we investigate extensive experiments on three popular benchmarks,
i.e.,
Human3.6M~\cite{h36m_pami},
3DPW~\cite{vonMarcard2018}
and
SURREAL \cite{varol17_surreal}.
Specifically,
Human3.6M is a popular motion capture dataset.
We use 5 subjects (S1, S5, S6, S7 and S8) for training and 2 subjects (S9 and S11) for testing.
%where we down-sample all the videos from 50fps to 10fps to reduce redundancy.
3DPW is an in-the-wild dataset with multiple actors occurred in the same image.
We use its official data split for training and testing.
SURREAL is a large-scale synthetic dataset with SMPL body annotations.
We directly evaluate its test set by our model pretrained on Human3.6M to show generalization capacity.
Moreover,
as suggested in the literature \cite{kanazawaHMR18,sun2019dsd-satn,Rong_2019_ICCV,Zimmermann_inproceedings},
we mainly use three protocols to measure accuracy of our generated mesh,
i.e.,
Mean Per Joint Position Error (MPJPE),
Procrustes Aligned MPJPE (PA-MPJPE),
Mean Per Vertex Error (MPVE).
We implement our PC-HMR frameworks as follows.
%,unless stated otherwise.
First,
we choose HMR~\cite{kanazawaHMR18} as our basic architecture.
Second,
for Human3.6m and SURREAL,
we use CPN \cite{hong2018cascaded} as 2D pose estimator,
and
VideoPose3D \cite{pavllo:videopose3d:2019} as 2D-to-3D pose lifter in our framework.
% Note that,
% since HMR 2D pose refers to 14 key points,
% we adapt VideoPose3D as a 14-key-point version (original 17-key-point).
For 3DPW,
we use PoseNet \cite{Moon_2019_ICCV_3DMPPE} as 3D pose estimator in our framework.
Finally,
we train all the modules separately,
according to their official codes with default hyperparameter and supervision settings.
Then,
we fine-tune the entire framework using 3D mesh as supervision,
where
we set 90/50 training epochs with 1024/128 mini-batch size for Human3.6M and 3DPW.
We set the learning rate as 0.001 for our calibration module and $1\times10^{-5}$ for other modules.
We implement our frameworks by PyTorch.
All the codes and models will be released afterwards.

\subsection{SOTA Comparison}
%In Table \ref{tab:SOTA},
%we compare our framework with the recent approaches on human mesh reconstruction.
%First,
%both serial and parallel PC-HMR frameworks significantly outperform HMR.
%This clearly demonstrates the importance of 3D pose guidance.
%Second,
%our parallel PC-HMR framework achieves the SOTA performance.
%For example,
%it outperforms DSD-SATN \cite{sun2019dsd-satn} that also leverages pose estimators for reconstruction.
%This shows our PC-HMR framework can use human pose in a more effective manner to boost mesh recovery.
%Additionally,
%for most metrics and benchmarks,
%our parallel PC-HMR outperforms SPIN \cite{fittingintheloop} that plugs iterative SMPLify optimizer into HMR for further calibrating mesh parameters.
%It shows our design is a more preferable calibration manner than traditional optimization.

In Table \ref{tab:SOTA},
our parallel PC-HMR framework achieves the SOTA performance,
e.g.,
for Human3.6M,
it outperforms \cite{sun2019dsd-satn,ChoiECCV2020} that also leverage pose estimators for reconstruction.
This shows our PC-HMR framework is a more effective manner to boost mesh recovery by human pose.
Additionally,
for most metrics and benchmarks,
our PC-HMR outperforms SPIN \cite{fittingintheloop} that plugs iterative SMPLify optimizer into HMR for further calibrating mesh parameters.
It shows our pose calibration is a more preferable calibration design than traditional optimization.

%We compare on Human3.6, 3DPW  and SURREAL. For both two-stream and pose encoder frameworks, we choose VideoPose3D \cite{pavllo:videopose3d:2019} as pose estimator and HMR \cite{kanazawaHMR18} as mesh generator to compare with previous approaches. Note that our frameworks are designed in plug-in style, which means the mesh and pose modules in both frameworks can be easily replaced with other 2D-3D pose lifter and mesh reconstruction methods. As shown in Table, our two-stream outperforms the state-of-the-art approaches including HMR, showing its superiority. Besides that, our pose encoder framework also show significant improvements compared with baseline, HMR. Additionally, we show generalization capacity of our frameworks. Our approach significantly reduces the mesh reconstruction error (i.e., MPVE) on all three datasets. It indicates that, both frameworks are practical and effective for improving mesh reconstruction.

\begin{figure*}[t]
\centering
\includegraphics[width=0.88\textwidth]{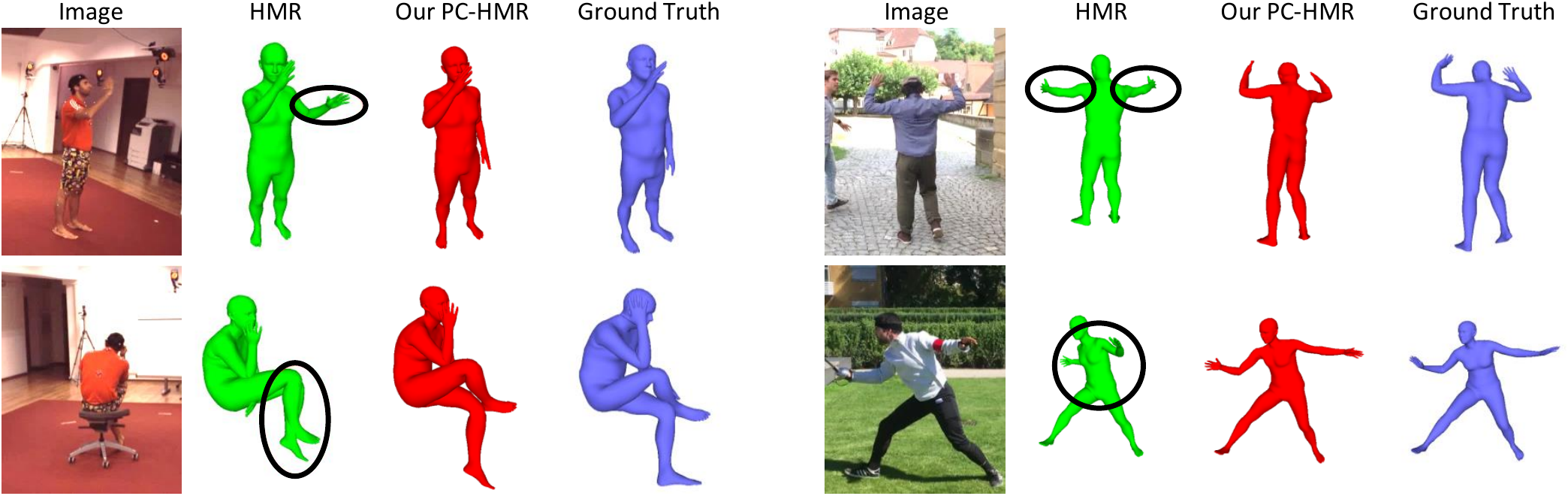}
\caption{Human Mesh Visualization. The left/right columns are respectively from Human3.6M/3DPW.}
\label{fig:vis}
\vspace{-0.3cm}
\end{figure*}
\begin{figure*}[t]
\centering
\includegraphics[width=0.93\textwidth]{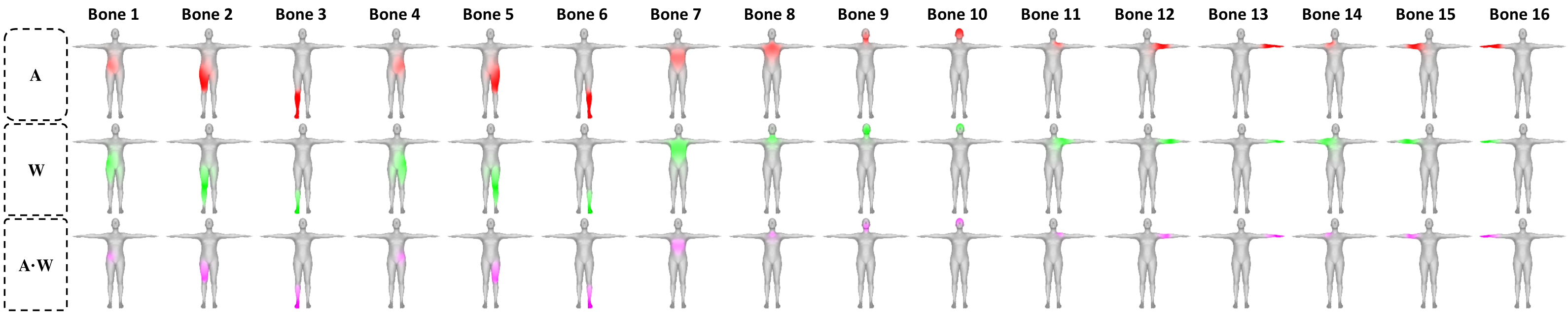}
\caption{Weight Visualization.
For bone $i$,
we visualize the trained
$\mathbf{A}(:, i)$,
$\mathbf{W}(:, i)$,
and $\mathbf{A}(:, i)\cdot\mathbf{W}(:, i)$ on the mesh,
where
the grey color indicates weights are close to zero,
while the bright color means weights are close to one.}
\label{fig:no_Wj}
\vspace{-0.5cm}
\end{figure*}

\begin{figure}[t]
    \centering
    \includegraphics[width=0.9\columnwidth]{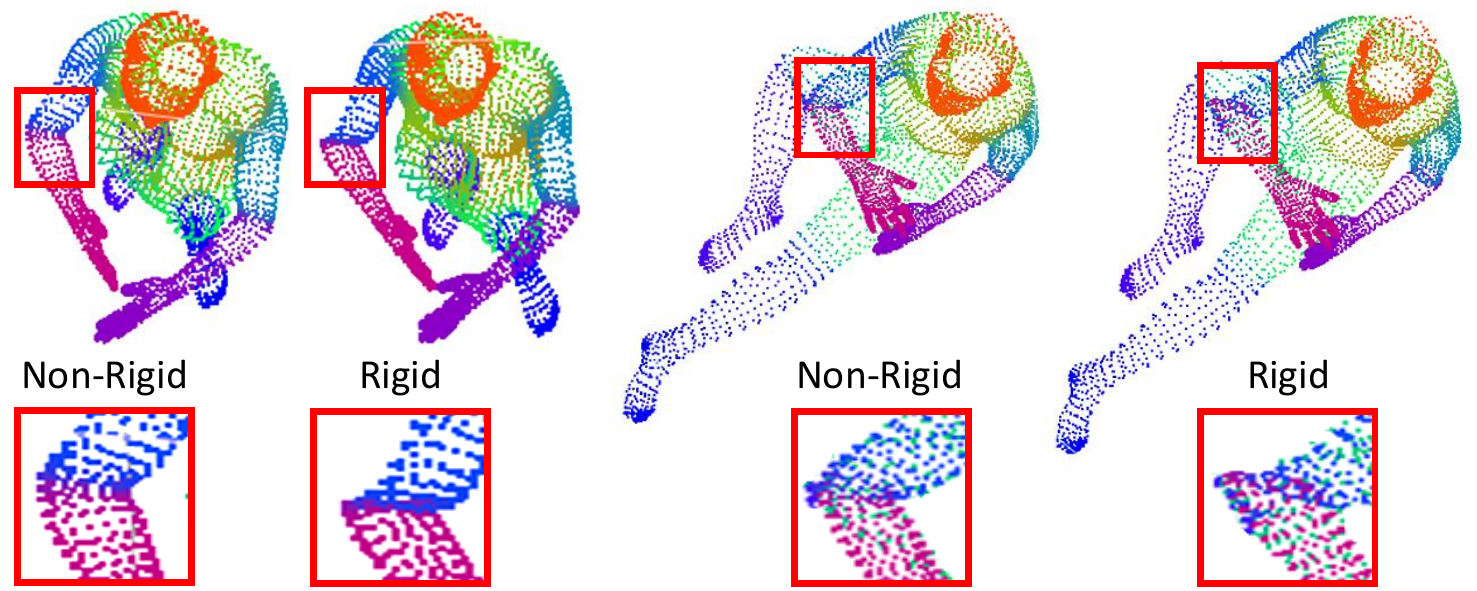}
    \caption{Non-Rigid vs. Rigid Transformation.
%The upper pink mesh results are using rigid transformation. The red circles show the elbow dislocation problem. The lower green ones are using non-rigid transformation.
%Obviously, human mesh with rigid transformation has significant dislocations around elbow joints, while our non-rigid transformation effectively reduces this problem.
}
\label{fig:nonrigid}
\vspace{-0.5cm}
\end{figure}

\subsection{Ablation Studies}
\label{sect:ablation}

\textbf{Generalization Capacity}.
We mainly use Human3.6M to further investigate properties of our PC-HMR frameworks.
In Table \ref{tab:plugin},
we examine generalization capacity of our frameworks,
via changing different mesh generators, 2D-to-3D pose lifters and 2D pose estimators.
First,
both serial and parallel frameworks outperform baselines,
no matter which types of mesh generators, pose lifters and estimators we use.
It shows the effectiveness of our pose calibration.
Second,
our proposed frameworks take tradeoffs between recovery error and computation cost,
i.e.,
Serial PC-HMR is lighter while Parallel PC-HMR is more accurate.
This provides more reasonable choices in practice,
depending on which factor is important in the deployment.
Third,
both frameworks can be straightforwardly used for video-based mesh reconstruction without any difficulties.
For example,
we use video-based HMMR \cite{learning3dhumandynamics} as mesh generator,
or
use video-based VideoPose3D \cite{pavllo:videopose3d:2019} as 2D-to-3D pose lifter in our frameworks.
We can see that,
video-based frameworks outperform image-based frameworks,
no matter which styles we use (serial or parallel).
It shows the importance of using temporal information.
Moreover,
VideoPose3D \cite{pavllo:videopose3d:2019} is more critical than HMMR \cite{learning3dhumandynamics} for temporal modeling,
e.g.,
when we choose VideoPose3D as pose lifter,
the performance would be comparable no matter which mesh generators we use (HMR or HMMR).
It indicates that,
it is more effective to introduce temporal modeling for 3D human pose,
without complex interference of learning mesh shape.
Finally,
we use 2D pose in the official code of VideoPose3D \cite{pavllo:videopose3d:2019},
where
2D pose estimator refers to CPN \cite{hong2018cascaded}.
Hence,
CPN \cite{hong2018cascaded} and VideoPose3D \cite{pavllo:videopose3d:2019} are more compatible to be a 3D pose estimator,
which leads to the best accuracy in our parallel framework.
For other parallel cases,
the better 2D pose estimator (e.g. HRNet) achieves a better performance of mesh recovery as expected.

%We mainly use Human3.6M to further investigate properties of our PC-HMR frameworks.
%In Table \ref{tab:plugin},
%we examine generalization capacity of our frameworks,
%via changing different mesh generators, 2D-to-3D pose lifters and 2D pose estimators.
%First,
%both serial and parallel frameworks significantly outperform baselines,
%no matter which types of mesh generators, pose lifters and estimators we use.
%It shows the effectiveness of our pose calibration.
%Second,
%our proposed frameworks take tradeoffs between recovery error and computation cost,
%i.e.,
%serial PC-HMR is more lighter while parallel PC-HMR is more accurate.
%This provides more reasonable choices in practice,
%depending on which factor is important in the deployment.
%Third,
%both frameworks can be straightforwardly used for video-based mesh reconstruction without any difficulties,
%e.g.,
%in the serial framework,
%we use HMMR \cite{humanMotionKZFM19} as video-based mesh generator,
%and VideoPose3D \cite{pavllo:videopose3d:2019} as video-based 2D-to-3D pose lifter.
%It is similar for the parallel case.
%Via such efficient adaptation,
%we can effectively address video-based human mesh recovery with high accuracy.
%All these facts show generalization capacity and flexibility of our frameworks.

\textbf{Detailed Designs}.
Since MPVE directly reflects recovery error of mesh surface,
we use it to evaluate the detailed designs.
Additionally,
we choose the parallel framework for this study,
due to its higher accuracy.
%In this case,
%we can clearly verify if our different designs still work within a highly-accurate framework.
%It is the similar case for serial framework.
\textbf{(I) w/o Non-Rigid}.
For comparison,
we delete the non-rigid term $\mathbf{W}_{j,i}\bm{\Delta}^{(i)}$ of Eq. (\ref{eq:mesh_trans2}) in our pose calibration module.
As shown in Table \ref{tab:calib},
the non-rigid one achieves a smaller mesh error,
showing its effectiveness.
\textbf{(II) w/o Fine-Tune}.
In our design,
we fine-tune the entire framework after training each module separately.
We delete this fine-tuning procedure for comparison.
As expected,
the setting of fine-tuning is better,
since our pose calibration module becomes compatible for mesh recovery via end-to-end learning.
Additionally,
our pose calibration module can still boost HMR,
even without fine-tuning.
It mainly thanks to geometry modeling in this module.
%which can reduce training difficulties in the data-driven framework.
\textbf{(III) PC-Template}.
%We investigate if our pose calibration module is necessary to be coupled with HMR mesh.
%Hence,
In our parallel framework,
we replace the HMR mesh stream by a template mesh,
and operate pose calibration module to correct this template mesh to be target mesh of the input image.
In Table \ref{tab:calib},
our PC-HMR outperforms PC-Template.
It shows that HMR provides more preferable shape in the mesh,
and thus it is reasonable to integrate our pose calibration module in HMR.
\textbf{(IV) GT 3D Pose}.
We use GT 3D pose for comparison.
As expected,
GT 3D pose can achieve a better performance.
But still,
our PC-HMR with estimated 3D pose significantly outperforms HMR.
%But still,
%comparing with the gap between PC-HMR (estimated) and PC-HMR (GT),
%the gap between HMR and our PC-HMR (estimated) is much larger.
%This clearly shows the effectiveness of our PC-HMR.
\textbf{(V) Extra Self-Rotation}.
There may exist a relative rotation around the bone itself.
Hence,
we use a two-layer MLP to estimate it,
where we take mesh parameters $\bm{\Theta}$ in HMR and $\bm{\Psi}$ in our pose calibration module as input.
Then we multiply the self rotation matrix with Eq. (\ref{eq:bone_trans02}) as final rotation matrix in our calibration module.
In Table \ref{tab:calib},
the results are comparable between our default and extra self-rotation setting.
For simplicity,
we use our default setting in the experiments.
\textbf{(VI) w/o Shape Compensation}.
To reduce detailed shape error,
we use a two-layer MLP to estimate 3D rotation of keypoints in the calibrated mesh,
with input of mesh parameters $\bm{\Theta}$ in HMR and $\bm{\Psi}$ in our pose calibration module.
Then we feed the estimated 3D rotation into pose blend shape function of SMPL as post-processing compensation in our framework.
In Table \ref{tab:calib},
the w/o shape compensation setting is slightly worse.
Hence,
we use our default setting.

\subsection{Visualization}

\textbf{Mesh Visualization}.
We visualize HMR (baseline) and our PC-HMR (parallel) in Fig. \ref{fig:vis}.
%Due to self-occlusion in Human3.6M,
%we show 3D mesh from alternative viewpoint.
As expected,
our PC-HMR can generate 3D mesh with more reliable pose,
even for occluded (e.g., Human3.6M) or in-the-wild (e.g., 3DPW) scenarios.
It indicates the effectiveness of our model.

\textbf{Weight Visualization}.
We visualize the trained weights of our pose calibration module,
i.e.,
$\mathbf{A}$ in Eq. (\ref{eq:mesh_trans3}),
$\mathbf{W}$ in Eq. (\ref{eq:mesh_trans2})
and
$\mathbf{A}\cdot\mathbf{W}$.
The parallel framework is used as illustration in Fig. \ref{fig:no_Wj}.
First,
$\mathbf{A}(:, i)$ controls the importance of bone $i$ when obtaining each final vertex in the calibrated mesh.
Hence,
bone $i$ has more contribution on the vertices around it.
% \textcolor{red}{Similarly, $\mathbf{W}(:, i)$ controls the importance of joints $i$ when obtaining each final vertex in the calibrated mesh.}
Second,
according to Eq. (\ref{eq:mesh_trans2})-(\ref{eq:mesh_trans3}),
$\mathbf{A}(:, i)\cdot\mathbf{W}(:, i)$ controls the proportion of non-rigid term $\bm{\Delta}^{(i)}$ in bone $i$.
Via further learning $\mathbf{W}(:, i)$,
we can see that $\mathbf{A}(:, i)\cdot\mathbf{W}(:, i)$ is gradually highlighted around the child joint of each bone,
in order to proportionally adjust bone length in the calibrated mesh.
Both facts allow us to calibrate HMR mesh smoothly and reasonably with non-rigid transformation.

\textbf{Calibration Module}.
We further show non-rigid transformation in our pose calibration module.
We use parallel framework as illustration in Fig. \ref{fig:nonrigid}.
As expected,
the generated mesh with our non-rigid transformation is much more natural,
without significant misplacement.

\section{Conclusion}
In this paper,
we design two generic plug-and-play PC-HMR frameworks to calibrate human mesh with explicit guidance of 3D pose.
They leverage a non-rigid pose calibration module to couple HMR mesh generators and 3D pose estimators in the serial or parallel manner,
so that they can be flexibly applied for image/video-based mesh recovery,
and have no requirement of 3D pose annotations in the testing.
The extensive experiments on popular benchmarks show our frameworks significantly boost recovery performance.

\section*{Acknowledgments}
This work is partially supported by Guangdong Special Support Program (2016TX03X276), and National Natural Science Foundation of China (61876176,U1713208), Shenzhen Basic Research Program (CXB201104220032A), the Joint Lab of CAS-HK. 
{\small
\bibliographystyle{aaai21}
\bibliography{bibFile}
}
\end{document}